\documentclass{article}

\usepackage{microtype}
\usepackage{graphicx}
\usepackage{booktabs} 
\usepackage{pgfplots}
\usepackage{pgfplotstable}
\usepackage{subfig}
\usepackage{multirow}
\usepackage{ulem}
\usepackage{hyperref}
\usepackage{enumitem}
\setlist{topsep=0pt}

\usepackage[accepted]{icml2025}

\usepackage{amsmath}
\usepackage{amssymb}
\usepackage{mathtools}
\usepackage{amsthm}

\usepackage[capitalize,noabbrev]{cleveref}

\newcommand{\name}{\text{KAN-AD~}}
\newcommand{\namenospace}{\text{KAN-AD}}
\newcommand{\mname}{\text{function deconstruction}}
\newcommand{\mn}{\text{FD}}
\newcommand{\codeurl}{\url{https://github.com/CSTCloudOps/KAN-AD}}

\theoremstyle{plain}

\theoremstyle{definition}

\theoremstyle{remark}

\usepackage[textsize=tiny]{todonotes}

\usepackage[final]{changes}

\newcommand{\acceptdeleted}[1]{}     
\definechangesauthor[name={Changhua Pei}, color=blue]{pch}
\definechangesauthor[name={Quan Zhou}, color=orange]{zq}
\newcommand{\pchadd}[1]{\added[id=pch]{#1}}

\newcommand{\zq}[1]{\added[id=zq]{#1}}

\icmltitlerunning{KAN-AD: Time Series Anomaly Detection with Kolmogorov–Arnold Networks}

\begin{document}

\twocolumn[
\icmltitle{KAN-AD: Time Series Anomaly Detection with Kolmogorov–Arnold Networks}




\begin{icmlauthorlist}
\icmlauthor{Quan Zhou}{cnic,ucas}
\icmlauthor{Changhua Pei}{cnic,hias}
\icmlauthor{Fei Sun}{ict}
\icmlauthor{Jing Han}{zte}
\icmlauthor{Zhengwei Gao}{zte}
\icmlauthor{Haiming Zhang}{cnic}
\icmlauthor{Gaogang Xie}{cnic}
\icmlauthor{Dan Pei}{thu}
\icmlauthor{Jianhui Li}{nju,cnic}
\end{icmlauthorlist}

\icmlaffiliation{cnic}{Computer Network Information Center, Chinese Academy of Sciences}
\icmlaffiliation{ucas}{University of the Chinese Academy of Sciences
}
\icmlaffiliation{hias}{Hangzhou Institute for Advanced Study, University of the Chinese Academy of Sciences}
\icmlaffiliation{ict}{Institute of Computing Technology, Chinese Academy of Sciences}
\icmlaffiliation{zte}{ZTE}
\icmlaffiliation{nju}{School of Frontier Sciences, Nanjing University}

\icmlcorrespondingauthor{Jianhui Li}{lijh@nju.edu.cn}
\icmlcorrespondingauthor{Changhua Pei}{chpei@cnic.cn}
\icmlaffiliation{thu}{Department of Computer Science and Technology, Tsinghua University}

\icmlkeywords{Machine Learning, Time Series}

\vskip 0.3in
]
\printAffiliationsAndNotice{}




\begin{abstract}
Time series anomaly detection (TSAD) underpins real-time monitoring in cloud services and web systems, allowing \replaced[id=zq]{rapid}{prompt} identification of anomalies to prevent costly failures.
Most TSAD methods driven by forecasting models\deleted[id=zq]{, trained with MSE loss,} tend to overfit by emphasizing minor fluctuations.
{Our analysis reveals that effective TSAD should focus on modeling ``normal'' behavior through smooth local patterns\deleted[id=zq]{, rather than attempting to capture detailed fluctuations}.}
To achieve this, we reformulate \deleted[id=zq]{normal} time series modeling as approximating the series with smooth univariate functions\replaced[id=zq]{. The local smoothness of each univariate function ensures that the fitted time series remains resilient against local disturbances. However, a direct KAN implementation proves susceptible to these disturbances due to the inherently localized characteristics of B-spline functions.}{, and adopt the Kolmogorov–Arnold Networks (KAN) framework as our backbone.
However, standard KAN suffers from performance drops in real-world TSAD due to its \deleted[id=zq]{default} B-spline design.}
We thus propose \textbf{KAN-AD}, replacing B-splines with truncated Fourier expansions and introducing a novel lightweight learning mechanism that emphasizes global \replaced[id=zq]{patterns}{regularity} while staying robust to local disturbances.
On four popular TSAD benchmarks, KAN-AD achieves an average 15\% improvement in detection accuracy (with peaks exceeding 27\%) over state-of-the-art baselines.
Remarkably, it requires fewer than 1,000 trainable parameters, resulting in a 50\% faster inference speed compared to the original KAN, demonstrating the approach’s efficiency and practical viability.
\end{abstract}

\section{Introduction} \label{intro}

Time Series Anomaly Detection (TSAD) serves as a critical component in modern IT infrastructure~\cite{ict2, ict4} and manufacturing systems~\cite{mi3, mi4}, enabling rapid identification of potential anomalies and providing sufficient clues for fault localization~\cite{triad, vqrae}.
The emergence of deep learning-based forecasting approaches~\cite{anomalytransformer, timesnet, ofa} have superseded traditional rule-based methods~\cite{lof, spot}, establishing new state-of-the-art performance through their capacity to {fit} historical data and detect anomalies via prediction-observation comparisons. 

\begin{figure}
    \setlength{\abovecaptionskip}{2pt}
    \centering
    \includegraphics[width=1\linewidth]{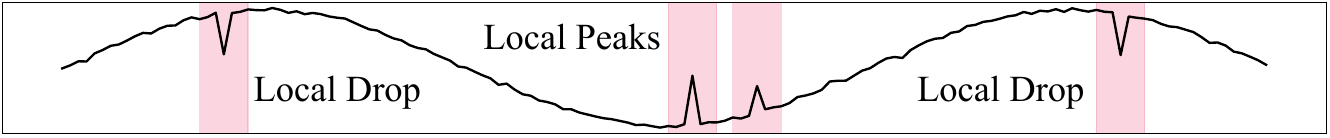}
    \caption{Illustration of local drops and peaks.}
    \label{fig:anomaly}
    \vspace{-1em}
\end{figure}

\begin{figure}
    \setlength{\abovecaptionskip}{2pt}
    \centering
    \includegraphics[width=1\linewidth]{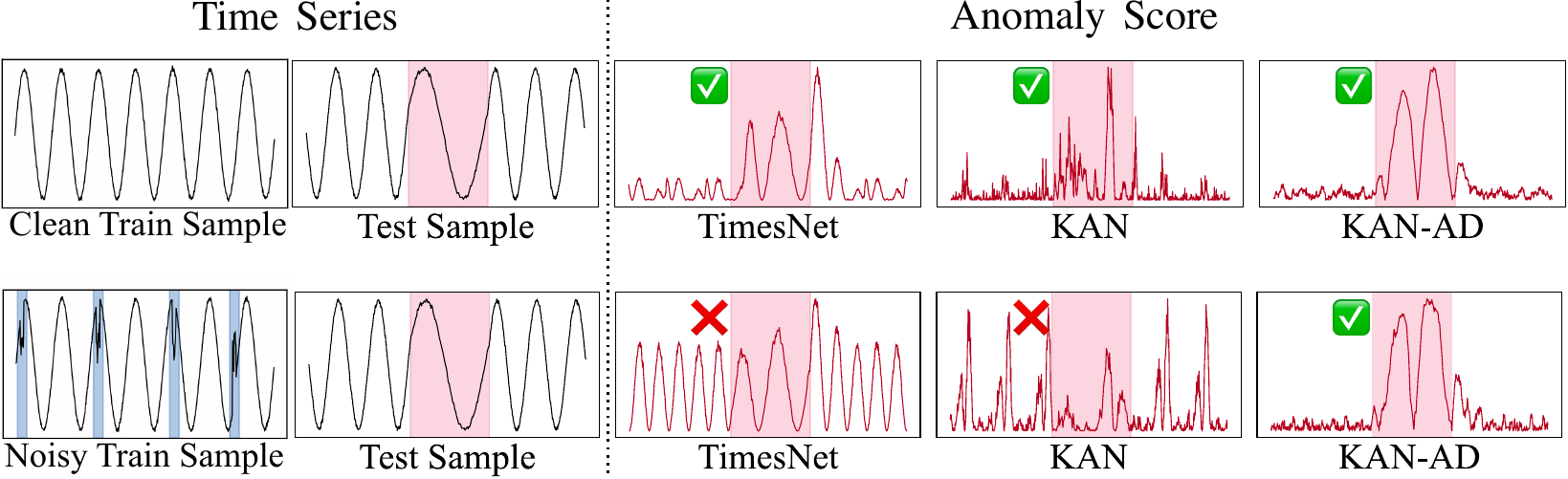}
    \caption{Comparison of anomaly detection performance. Top: All methods successfully detect anomalies when trained on clean data (black curve, anomalous segments in pink). Bottom: TimesNet and KAN fail to detect anomalies when trained on noisy data. Blue markers indicate local drops and peaks; red curve shows anomaly scores.}
    \label{fig:noise-sensitive}
    \vspace{-1.5em}
\end{figure}

However, the effectiveness of the forecasting-based approach declines when encountering time series with localized disturbances.
As illustrated in~\Cref{fig:anomaly}, time series data frequently exhibit local peaks and drops that can significantly impact model learning. \replaced[id=zq]{Existing}{Contemporary} deep learning methods~\cite{tuli2022tranad, timesnet} often overfit to these local disturbances, compromising their ability to {detect anomalies} effectively. From the third column of~\Cref{fig:noise-sensitive}, we can observe that compared to training with clean data, TimesNet~\cite{timesnet} trained on noisy data fails to detect anomalies in the samples.

\replaced[id=zq]{Our experimental analysis reveals that forecasting-based TSAD methods suffer performance degradation by attempting to model every detailed patterns in raw time series data. While these methods aim to identify anomalies through comparison with predicted behavior, such detailed modeling proves unnecessary and potentially detrimental, especially given that real-world time series typically contain various forms of anomalies and irrelevant disturbances, presenting two significant challenges: }{We assert that performance degradation of forecasting methods primarily results from their attempt to fit the raw time series. This is because the primary aim of TSAD is to identify anomalies by forecasting the \textbf{normal} pattern and subsequently comparing it to the actual pattern that occurred. It is unnecessary to learn every detail of the raw time series, as real-world time series often have useless disturbances. }\pchadd{firstly, \replaced[id=zq]{the difficulty in establishing a universal criterion for filtering these disturbances}{creating a single rule to eliminate these disturbances is difficult}, and secondly, developing another model to ensure the forecasting model's input is free of local disturbances is resource-intensive. \replaced[id=zq]{Given these inherent limitations in both filtering-based and dual-modeling approaches, researchers have explored VAE-based approaches to address the challenge of local disturbance mitigation. VAE-based approaches~\cite{donut,fcvae} assume that normal patterns in time series cluster in a low-dimensional latent space and can be effectively reconstructed, thereby overcoming interference from data perturbations. Nevertheless, as demonstrated in FCVAE~\cite{fcvae}, VAE-based approaches struggle with underfitting, which impairs their ability to reconstruct the original time series and limits their effectiveness.}{VAE-based approaches aim to explicitly capture normal patterns by transforming high-dimensional time series into a low-dimensional latent space, assuming that normal data are densely clustered in this reduced space and can be reconstructed back effectively. Nevertheless, as indicated in FCVAE, VAE-based approaches struggle with underfitting, which hinders their ability to adequately represent the original shapelet of time series and limits their effectiveness.}}

\replaced[id=zq]{To mitigate local disturbances, we reformulate TSAD by approximating time series using smooth univariate functions, building on the theoretical foundation that {normal sequences exhibit greater local smoothness than abnormal ones~\cite{anomalytransformer}.}}{\pchadd{To avoid the effects of local disturbances when learning the normal pattern, we reformulate the TSAD task's modeling of the ``normal'' pattern by approximating the series using smooth univariate functions. This approach stems from the observation that normal time series exhibit greater local smoothness compared to abnormal sequences, a finding also supported by AnoTransformer.}} To achieve \replaced[id=zq]{this}{the above} formulation, 
Kolmogorov-Arnold Networks (KANs)~\cite{kan} offer a promising direction by decomposing complex objectives into combinations of learnable {univariate} functions based on the Kolmogorov-Arnold representation theorem~\cite{ka2}. This decomposition approach has shown remarkable effectiveness in various domains~\cite{kaneffective1, kaneffective2}. However, direct application of KAN to TSAD presents significant challenges. From the \pchadd{fourth column in the} upper part of~\Cref{fig:noise-sensitive}, it can be observed that models trained on clean training samples can detect anomalies in the test samples. But we find that \deleted[id=zq]{even} KAN fails to detect anomalies when the training samples contain noisy samples \deleted[id=zq]{\pchadd{when focusing on the bottom row of~\Cref{fig:noise-sensitive}}}. {The main reason is that, although KAN can specify {univariate} functions, \pchadd{\textit{i.e.,} B-spine function,} these functions are not specifically designed for time series and can still overfit local features, failing to completely eliminate the impact of local peaks or drops}.

To address these challenges, we propose \namenospace, adopting KAN as our backbone. By considering the characteristics of time series, we redesign KAN in three aspects. First, we replace the B-spine function with Fourier series. \replaced[id=zq]{Fourier series have local smoothness compared to spline functions, while their natural periodicity allows for better modeling of global patterns~\cite{fourierglobal1,fourierglobal2}.}{Fourier Series represent a periodic function as a sum of sine and cosine terms, which aligns well with the periodic properties of time series and have local smoothness at the same time. Compared to spline functions in KAN, Fourier series naturally extract frequency domain information from time series, enabling better modeling of global patterns~\cite{fourierglobal1,fourierglobal2}} Second, as the Fourier series contains unlimited terms which is computation intensive, we only use the first $N$ terms of Fourier series. To overcome the limitation that the first N terms of Fourier series can only model periodic no smaller than $\frac{1}{N}$, we designed an alternative index-based univariate function to capture the fine-scale periodic missing from the first N terms. \replaced[id=zq]{Third, we incorporated differencing to isolate time series trend effects on coefficient estimation, leading to improved modeling accuracy through more precise coefficients.}{Third, we incorporate a differencing mechanism to decouple the negative impact of trend components in TSAD. }

Our comprehensive evaluation demonstrates that KAN-AD achieves 15\% higher F1 accuracy while being 50\% faster than the original KAN architecture. Our code is publicly available at {\codeurl}. Our contributions are as follows:
\begin{itemize}
    \setlength{\itemsep}{-0.5em}
    \item We reformulate the problem to assist deep learning-based forecasting models for time series anomaly detection (TSAD) tasks by minimizing overfitting to local perturbations.
    \item We introduce \namenospace, an innovative TSAD approach. KAN-AD, built meticulously on the KAN backbone, \replaced[id=zq]{exhibits substantial improvements in both detection precision and inference efficiency}{surpasses KAN in detection precision and offers quicker inference}.
    \item We performed comprehensive experiments on four publicly available datasets, verifying the effectiveness and efficiency against state-of-the-art TSAD benchmarks.
\end{itemize}
\section{Preliminaries and Problem Formulation} \label{preliminary}
\subsection{Problem Statement} \label{statement}
This paper primarily addresses the issue of anomaly detection in single time series curves, also known as univariate time series (UTS). To elaborate on the problem more comprehensively, consider the following UTS observational data: $ x_{0:t} = \{ x_0, x_1, x_2, \dots, x_t \} $ and anomaly labels $ C = \{ c_0, c_1, c_2, \dots, c_{t}\} $, where $ x_t \in \mathbb{R} $, $ c \in \{0, 1\} $, and $ t \in \mathbb{N} $. Here, $x_{0:t}$ represents the entire observed time series, and $C$ denotes the temporal anomaly labels. 

\textit{Given a UTS $x=[x_0,x_1,x_2,\dots,x_t]$, the objective of UTS anomaly detection is to utilize the data $[x_0,x_1,\dots,x_i]$ preceding each point $x_i$ to predict $c_i$}.

\subsection{Kolmogorov–Arnold Networks}

\subsubsection{Theoretical Foundation} The Kolmogorov–Arnold representation theorem demonstrates that any multivariate continuous function can be decomposed into a finite sum of univariate functions, as shown in~\Cref{agn:ka}, where $\varphi_{q,p}$ are univariate functions that map each input variable $x_p$, and $\Phi_q$ are continuous functions.
\begin{align}
f(x_1,x_2,\dots,x_n)&=\sum_{q=1}^{2n+1}\Phi_q(\sum_{p=1}^n\varphi_{q,p}(x_p)) \label{agn:ka} \\
\text{KAN}(x) & = (\Phi_{L-1} \circ \Phi_{L-2} \circ \dots \circ \Phi_0)(x) \label{agn:kan}
\end{align}

\subsubsection{Network Architecture and Function Representation}
KAN consists of a series of interconnected univariate sub-networks, each responsible for learning distinct features of the data. Unlike traditional multi-layer perceptrons (MLPs), which employ fixed activation functions at each node, KAN replaces each weight parameter with a univariate function. The resulting functional form for deeper KAN can be expressed as~\Cref{agn:kan}, where each $\Phi_l$ represents a layer of univariate functions applied to the input or intermediate outputs. 
The vanilla KAN~\cite{kan} implements these univariate functions using B-splines~\cite{bspline}, which provide localized function approximation capabilities. However, this localization property presents a notable consideration in anomaly detection contexts. Since anomalous patterns typically manifest as localized features~\cite{anomalytransformer}, B-splines may inadvertently fit these outliers, potentially compromising model accuracy.

\section{Methodology} \label{methodology}
\begin{figure*}[t]
    \centering
    \subfloat[Illustration of learning components in KAN and \namenospace. \name learns the coefficients on edges with fixed univariate functions, and performs weighted sum operations on nodes. Blue lines indicate edges with weights.\label{fig:compare}]{
        \includegraphics[width=0.3\linewidth]{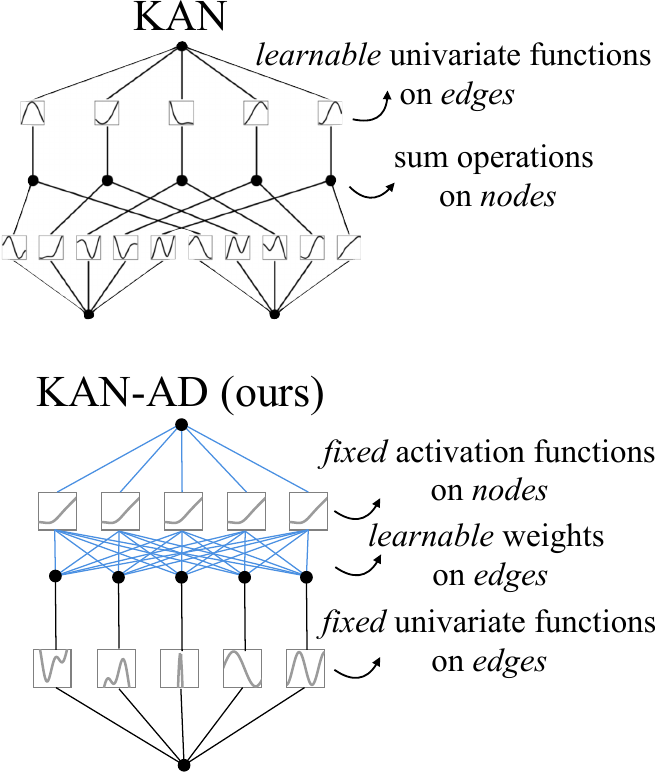}
    }
    \hspace{0.75em} 
    \subfloat[Illustration of the \name process using a sliding window approach. During the mapping phase, raw time windows are transformed into multiple univariate functions. In the reducing phase, a one-dimensional convolutional kernel learns coefficients for these univariate functions, aggregating them into a normal pattern for the current time window. In the projection phase, a linear layer predicts future normal patterns.\label{fig:model-arch}]{
        \includegraphics[width=0.65\linewidth]{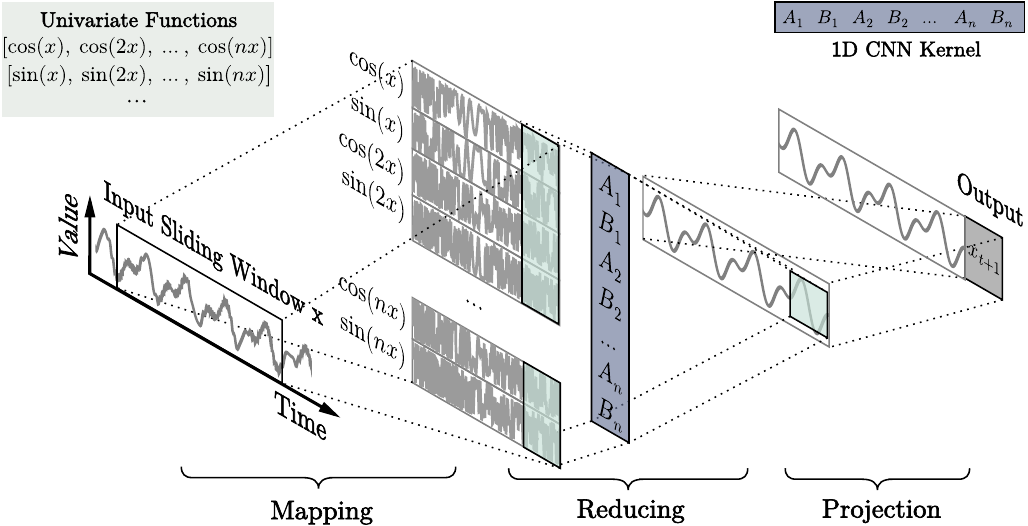}
    }
    \caption{Illustration of \namenospace.}
    \label{fig:main-figure}
    \vspace{-1em}
\end{figure*}
\zq{The core challenge in time series anomaly detection (TSAD) lies in establishing accurate normal patterns while maintaining robustness to local disturbances~\cite{interfusion}. Traditional approaches that directly predict based on historical data inevitably incorporate local noise into their learned patterns. Building on the observation that normal sequences exhibit greater smoothness than abnormal ones, we propose \namenospace, a novel anomaly detection framework that leverages this smoothing feature to identify anomalies in complex time series data.}

\subsection{Design of \namenospace}
\zq{The pipeline of \name consists of three main stages: \textbf{mapping}, \textbf{reducing}, and \textbf{projection}. In the mapping phase, we decompose the input time window into multiple univariate functions. The reducing phase then combines these functions through learned coefficients to reconstruct the ``normal'' pattern. Finally, the projection phase leverages this pattern to predict future behavior, enabling anomaly detection through comparison with real-time observations.}
\begin{align}
f(x_{0:i}) = A_0 & + \underbrace{\sum_{n=1}^{N}{\left(A_n \cos(nx_{0:i}) + B_n \sin(nx_{0:i})\right)}}_{g(x_{0:i})} + \epsilon
\label{agn:fourier-formula}
\end{align}
\vspace{-1em}
\begin{align}
&\mathbf{H} = \mathsf{Stack}(
\cos(x_{0:i}),\sin(x_{0:i}),\dots,\cos(nx_{0:i}),\sin(nx_{0:i})
) \notag \\
&\hspace{2em}\mathbf{\Theta} (x_{0:i}) = \begin{bmatrix}A_1,B_1,A_2,B_2,\dots,A_n,B_n\end{bmatrix} \nonumber \\
&\hspace{3.7em} x'_{0:i} = A_0 + \mathbf{\Theta}(x_{0:i}) \times \mathbf{H}
\label{agn:fan-formula}
\end{align}

\zq{Formally, we employ Fourier series for normal pattern representation, motivated by two key advantages over alternative approaches such as B-spline functions. First, the constituent sine and cosine functions exhibit superior local smoothness, avoiding the potential overfitting to local noise. Second, Fourier series naturally capture global patterns, particularly excelling at modeling periodic behaviors in time series.
Following this motivation, we introduce the \mname~(FD) mechanism, where $f$, the mapping between the historical window $x_{0:i}$ and its next behavior $x_{i+1}$, can be expanded as shown in~\Cref{agn:fourier-formula}. The normal pattern can be represented by the finite $N$ terms of the series~\cite{ka2}, denoted as $g(x)$, while the terms beyond $N$ capture the stochastic observational noise $\epsilon$. The normal pattern $x'_{0:i}$ can then be expressed as in~\Cref{agn:fan-formula}, where $\mathbf{H}$ denotes the univariate function matrix. \replaced[id=zq]{This decomposition combined with learnable coefficients filters out potential noise and significantly simplifies the construction of normal patterns.}{This decomposition transforms potentially noisy time series into locally smooth normal patterns, significantly simplifying the subsequent prediction task.}} 

\subsection{Mapping Phase}
As shown in~\Cref{fig:model-arch}, the primary purpose of the mapping phase is to transform the original time series signal \( x_{0:i} \in \mathbb{R}^{T} \) into multiple new sets of values \( x_{0:i} \in \mathbb{R}^{T \times (N+N)} \) through a series of univariate functions. Here, $T$ is the size of the sliding window. The first \( N \) represents the number of sine series univariate functions, and the other \( N \) represents the number of cosine series univariate functions. The detailed calculation method is shown in~\Cref{agn:fourier-formula}. Notably, besides the univariate function terms, an $A_0$ term representing the average value within the sliding window is also present, which varies across different windows. To mitigate the impact of fluctuating $A_0$ on coefficient fitting, a constant term elimination module is employed.

\textbf{Constant Term Elimination}: In Fourier series,  $A_0$ represents the mean value of the function. Although normalization ensures that the entire time series has a mean of zero, individual time windows may still exhibit significant fluctuations in their means due to the presence of a trend. These variations in the constant term ultimately affect the model's accurate estimation of Fourier coefficients, leading to biases in the construction of the normal pattern. 

To mitigate the impact of mean fluctuations on the model's approximation of normal time series patterns, we employ \textit{first-order differencing} during data preprocessing to minimize the residual trend component in the data and subsequently renormalize the differenced data. This strategy allows the model to focus on estimating Fourier coefficients $A_{1:n}$ and $B_{1:n}$, thereby avoiding the need to learn frequently changing constant terms. After this differential strategy, the normal pattern \( x'_{0:i} \) can be expressed as $x'_{0:i} \sim \mathbf{\Theta}(x_{0:i}) \times \mathbf{H}$

\textbf{Periodic-Enhanced \namenospace}:  \replaced[id=zq]{Fourier series of finite $N$ terms cannot model a period smaller than $\frac{1}{N}$, which limits \namenospace's ability to express time series containing more subtle periods.}{Traditional Fourier series decomposition represents periodic time series through fixed periodic components, which lacks adaptability to varying temporal dependencies in complex time series. }

To address this limitation and enhance the model's ability to capture periodic patterns in time series, we introduce additional univariate functions with different periods. Specifically, we incorporate trigonometric components $\cos(\frac{2\pi ni}{T})$ and $\sin(\frac{2\pi ni}{T})$ where $i$ denotes the window index, with coefficients learned through one-dimensional convolution networks. Our implementation utilizes three complementary univariate functions shown in~\Cref{agn:feature-set}: the raw time variable $X$, the Fourier series $S_n$, and the sine-cosine wave $P_n$. This integration of multi-periodic univariate functions enhances \name's capacity to model temporal patterns.
\begin{align}
    \mathrm{X}\  &= x_{0:i} \notag \\ 
    \mathrm{S_n} &= \{\sin(nx_{0:i}), \cos(nx_{0:i})\} \label{agn:feature-set} \\
    \mathrm{P_n} &= \{\sin(\frac{2\pi ni}{T}), \cos(\frac{2\pi ni}{T})\} \notag
\end{align}

\subsection{Reducing Phase}



Another challenge in real-world time series anomaly detection is the high computational cost. Existing methods often sacrifice efficiency for accuracy, making them impractical in resource-constrained or large-scale settings. \replaced[id=zq]{The \mname~(\mn) mechanism addresses this challenge by transforming the modeling of normal patterns into a weighted combination of univariate functions. This transformation substantially reduces the model's parameter quantity - instead of requiring numerous parameters for fine-grained feature modeling, \mn~mechanism achieves efficient representation through estimating coefficients of a small number of univariate functions.}{The \mname~(\mn) mechanism addresses this challenge by transforming the modeling of normal patterns from fine-grained weight adjustments to coefficient estimation of univariate functions. This transformation significantly reduces the model's parameter count since the number of univariate functions is substantially smaller than the time window length.}

\begin{align}
&\mathbf{H}^{(0)} {=} \mathsf{Stack}(X, S_1, P_1, \dots, S_n, P_n), \forall n {\in} \left [1,\dots,N \right]
\label{agn:h0} \\
&\mathbf{H}^{(l)}  = \mathsf{CNN}(\mathsf{CNN}(\mathbf{H}^{(l-1)})) \quad \forall l\in [1,2,\dots, L]\label{agn:2layers} \\ 
&\hspace{1.5em}\mathsf{Conv}(\mathbf{H})  =  \sum_{c=1}^{2N} \sum_{m=0}^2 W_c[m]\cdot \mathbf{H}_c[i+m-1] \label{agn:conv} \\
&\hspace{1.5em}\mathsf{CNN}(\mathbf{H}) = \mathsf{GELU}(\mathsf{BN}(\mathsf{Conv}(\mathbf{H}))) \label{agn:single-layer}
\end{align}

To effectively estimate these univariate function coefficients, we employ a stacked one-dimensional convolutional neural network (1D CNN). This architecture choice is motivated by two key factors: 1D CNNs excel at sequence modeling through temporal dimension traversal, while their convolutional kernels naturally capture the diverse features introduced by the \mn~mechanism. As shown in~\Cref{agn:h0}, \name first constructs a univariate function matrix $\mathbf{H}^{(0)}$ by combining the required functions for a given time window. This matrix is then processed through multiple stacked 1D convolutional layers with a kernel size of 3, progressively approximating the normal pattern through coefficient learning, as expressed in~\Cref{agn:2layers}. Here, $L$ denotes the number of CNN blocks, with the network $\mathsf{CNN}(\mathbf{H})$ and convolution operation $\mathsf{Conv}(\mathbf{H})$ defined in~\Cref{agn:single-layer,agn:conv}.
The convolution operation in~\Cref{agn:conv} applies a kernel $W_c$ to each channel $\mathbf{H}_c$, where indices $m$ and $t$ represent positions within the convolutional kernel and time window, respectively.

To ensure training stability and reduce internal covariate shift, we apply batch normalization~\cite{batchnormalization} after each convolutional layer (\Cref{agn:single-layer}), followed by Gaussian Error Linear Units (GELUs)~\cite{gelu} for activation. The final stage of the reducing phase employs a residual connection~\cite{resnet} between the hidden state $\mathbf{H}^{(L)}$ and the original input $\mathbf{H}^{(0)}$ to maintain numerical stability, as shown in~\Cref{agn:res}. Finally, a 1-width convolutional kernel reduces the dimensionality of $\mathbf{H}^{(L)'}$ to generate the normal pattern approximation $x'_{0:i}$ within the current time window:
\begin{align}
    \mathbf{H}^{(L)'} =\ & \mathbf{H}^{(L)} + \mathbf{H}^{(0)} \label{agn:res}\\ 
    x'_{0:i} =\ & \mathsf{GELU}(\mathsf{BN}(\mathsf{DownConv}(\mathbf{H}^{(L)'})) \label{agn:x'} 
\end{align}
Here, $\mathsf{DownConv}(\mathbf{H}) =\sum_{c=1}^{2N} W_c\cdot\mathbf{H}_c[i]$ denotes the convolution operation for reducing dimensions.



\subsection{Projection Phase}
After obtaining the current window's normal mode approximation $x'_{0:i}$, we predict the future behavior $x_{i+1}$ through a linear layer, leveraging \namenospace's accurate approximation capability:
\begin{equation}
x_{i+1} = W \cdot x'_{0:i} + b
\end{equation}
where $W$ and $b$ denote the weight matrix and bias term of the linear layer.
\section{Evaluation} \label{evaluation}

In this section, we conduct comprehensive experiments primarily aimed at answering the following research questions.

\noindent \textbf{RQ1}: How does \name compare to state-of-the-art anomaly detection methods in performance and efficiency?
\noindent \textbf{RQ2}: How sensitive is \name to hyperparameters?
\noindent \textbf{RQ3}: How effective is each design choice in \namenospace?
\noindent \textbf{RQ4}: How sensitive is \name to anomalies in the training data?

In addition, we also evaluate our method on a multivariate time series anomaly detection dataset to demonstrate the application potential of \name in more scenarios.

\subsection{Experimental settings} \label{eval:settings}


\subsubsection{Dataset} We evaluate \name on four publicly available UTS datasets: KPI~\cite{KPI}, TODS~\cite{suite_tods}, WSD~\cite{WSD}, and UCR~\cite{UCR}. Dataset characteristics are summarized in \Cref{tab:dataset}, including curve counts, sizes, and anomaly rates. The anomaly interval length distributions, shown in \Cref{fig:dataset-cdf}, reveal that while most anomalies span less than 10 points, WSD and UCR contain extended anomaly segments exceeding 300 points, enabling comprehensive evaluation. Detailed dataset descriptions are provided in \Cref{ap:dataset}.


\subsubsection{Model training and inference}

We implement a systematic experimental protocol for both our method and baseline approaches. For each time series, we train dedicated \name models using consistent hyperparameters: batch size 1024, learning rate 0.01, and maximum 100 epochs. The validation strategy varies by dataset, with UCR reserving 20\% of training data and other datasets employing a 4:1:5 ratio for training, validation, and testing splits. To ensure fair comparison, we faithfully replicate all baseline methods following their original implementations and hyperparameter settings as specified in their respective papers. During inference, we standardize the batch size to 1 across all methods for comparable efficiency assessment. Results presented in~\Cref{tb:main} report means and standard deviations from five independent trials with different random seeds.




\subsubsection{Baselines} 
\begin{table}[t]
    \centering
    \caption{Dataset Statistics.}
    \resizebox{0.95\linewidth}{!}{
        \renewcommand\arraystretch{1.0}
        \setlength{\tabcolsep}{1.1mm}{
            \begin{tabular}{lrrrrr} 
\toprule
\textbf{Dataset} & \textbf{Curves} & \textbf{Train} & \textbf{Train Ano\%} & \textbf{Test} & \textbf{Test Ano\% } \\ 
\hline
KPI & 29 & 3,073,567 & 2.70\% & 3,073,556 & 1.85\% \\
TODS & 15 & 75,000 & 5.32\% & 75,000 & 6.38\% \\
WSD & 210 & 3,829,373 & 2.43\% & 3,829,537 & 0.76\% \\
UCR & 203 & 3,572,316 & 0.00\% & 7,782,539 & 0.47\% \\
\bottomrule
\end{tabular}

        }
    }
    \label{tab:dataset}
    \vspace{-1em}
\end{table}

\begin{table*}[t]
    \setlength{\abovecaptionskip}{0.5em}
    \setlength{\belowcaptionskip}{-1em}
    \caption{Performance comparison. Best scores are highlighted in bold, and second best scores are highlighted in bold and underlined. Metrics include $\mathsf{F1}$ (Best F1), $\mathsf{F1_e}$ (Event F1), $\mathsf{F1_d}$ (Delay F1),  $\mathsf{AUPRC}$ (area under the precision-recall curve) and $\mathsf{Avg\ F1_e}$ (average $\mathsf{F1_e}$ score across four datasets).}
    \centering
    \LARGE
    \resizebox{\linewidth}{!}{
        \renewcommand\arraystretch{1.0}
        \setlength{\tabcolsep}{0.3mm}{
            \begin{tabular}{lrrrr rrrr rrrr rrrr r} 
\toprule
 \multirow{2.5}{*}{Method} & \multicolumn{4}{c}{\textbf{KPI}} & \multicolumn{4}{c}{\textbf{TODS}}   & \multicolumn{4}{c}{\textbf{WSD}} & \multicolumn{4}{c}{\textbf{UCR}} & \multirow{2.5}{*}{$\mathsf{Avg\ F1_e}$} \\ \cmidrule(lr){2-5} \cmidrule(lr){6-9} \cmidrule(lr){10-13} \cmidrule(lr){14-17}
 & \multicolumn{1}{c}{ $\mathsf{F1}$} & \multicolumn{1}{c}{ $\mathsf{F1_{e}}$} & \multicolumn{1}{c}{ $\mathsf{F1_{d}}$} & \multicolumn{1}{l}{ $\mathsf{AUPRC}$} & \multicolumn{1}{c}{ $\mathsf{F1}$} & \multicolumn{1}{c}{ $\mathsf{F1_{e}}$} & \multicolumn{1}{c}{ $\mathsf{F1_{d}}$} & \multicolumn{1}{l}{ $\mathsf{AUPRC}$} & \multicolumn{1}{c}{ $\mathsf{F1}$} & \multicolumn{1}{c}{ $\mathsf{F1_{e}}$} & \multicolumn{1}{c}{ $\mathsf{F1_{d}}$} & \multicolumn{1}{l}{ $\mathsf{AUPRC}$} & \multicolumn{1}{c}{ $\mathsf{F1}$} & \multicolumn{1}{c}{ $\mathsf{F1_{e}}$} & \multicolumn{1}{c}{ $\mathsf{F1_{d}}$} & \multicolumn{1}{l}{ $\mathsf{AUPRC}$} &   \\
\midrule
SRCNN                                    & 0.4137                          & 0.0994                              & 0.2266                              & 0.3355                           & 0.6239                          & 0.1918                              & 0.4399                              & 0.6076                           & 0.4092                          & 0.1185                              & 0.1951                              & 0.3080                           & 0.5964                          & 0.1369                              & 0.1656                              & 0.5109                           & 0.1367~                                \\
SAND                                     & 0.2710                          & 0.0397                              & 0.1097                              & 0.2022                           & 0.5372                          & 0.1879                              & 0.5103                              & 0.5145                           & 0.1761                          & 0.0839                              & 0.1267                              & 0.1238                           & 0.7044                          & \uline{0.5108}                      & \uline{0.5116}                      & 0.6550                           & 0.2056~                                \\
AnoTrans                                       & 0.6103                          & 0.3020                              & 0.3623                              & 0.5676                           & 0.4875                          & 0.1915                              & 0.2918                              & 0.4148                           & 0.4348                          & 0.2311                              & 0.1517                              & 0.3527                           & 0.6135                          & 0.1696                              & 0.1084                              & 0.5458                           & 0.2236~                                \\
TranAD                                   & 0.7553                          & 0.5611                              & 0.6399                              & 0.7399                           & 0.5035                          & 0.2460                              & 0.3619                              & 0.4501                           & 0.7570                          & 0.6338                              & 0.4158                              & 0.7106                           & 0.5278                          & 0.1840                              & 0.1554                              & 0.4599                           & 0.4062~                                \\
SubLOF                                   & 0.7273                          & 0.2805                              & 0.4994                              & 0.7015                           & 0.7997                          & 0.4795                              & 0.7169                              & 0.7809                           & 0.8683                          & 0.6585                              & 0.4917                              & 0.8353                           & \uline{0.8468}                  & 0.4772                              & 0.4151                              & \uline{0.8001}                   & 0.4739~                                \\
TimesNet                                 & 0.8022                          & 0.6363                              & 0.6995                              & 0.8166                           & 0.6232                          & 0.3327                              & 0.4495                              & 0.6031                           & 0.9406                          & 0.8444                              & 0.6170                              & 0.9376                           & 0.5273                          & 0.1805                              & 0.1439                              & 0.4536                           & 0.4985~                                \\
FITS                                     & 0.9083                          & 0.6353                              & 0.8175                              & 0.9359                           & 0.7773                          & 0.5416                              & 0.6312                              & 0.7725                           & 0.9732                          & 0.8391                              & 0.6535                              & 0.9771                           & 0.6664                          & 0.2926                              & 0.2912                              & 0.5969                           & 0.5772~                                \\
OFA                                      & 0.8810                          & 0.6150                              & 0.7952                              & 0.9009                           & 0.6928                          & 0.5811                              & 0.5588                              & 0.7206                           & 0.9564                          & 0.8344                              & 0.6250                              & 0.9615                           & 0.6294                          & 0.3176                              & 0.1503                              & 0.5699                           & 0.5870~                                \\
FCVAE                                    & 0.9398                          & 0.7556                              & 0.8624                              & 0.9572                           & \uline{0.8652}                  & 0.6995                              & 0.7482                              & \uline{0.8798}                   & 0.9650                          & 0.8610                              & 0.6583                              & 0.9653                           & 0.7651                          & 0.3812                              & 0.2857                              & 0.7145                           & 0.6743~                                \\
LSTMAD                                   & 0.9376                          & 0.7742                              & \textbf{0.8782}                     & 0.9624                           & 0.8633                          & \uline{0.6981}                      & \uline{0.7655}                      & 0.8740                           & 0.9866                          & \textbf{0.9028}                     & \textbf{0.6743}                     & 0.9849                           & 0.7040                          & 0.3482                              & 0.3121                              & 0.6432                           & 0.6808~                                \\
KAN                                      & \uline{0.9411}                  & \uline{0.7816}                      & 0.8666                              & \uline{0.9664}                   & 0.8109                          & 0.6466                              & 0.7518                              & 0.8286                           & \uline{0.9879}                  & \uline{0.8939}                      & \uline{0.6650}                      & \textbf{0.9881}                  & 0.8016                          & 0.4120                              & 0.3971                              & 0.7489                           & \uline{0.6835}~                        \\
\hline
\multirow{2}{*}{KAN-AD} & \textbf{0.9442} & \textbf{0.7989} & \uline{0.8755} & \textbf{0.9693} & \textbf{0.9425} & \textbf{0.8940} & \textbf{0.8391} & \textbf{0.9716} & \textbf{0.9888} & 0.8927 & 0.6623 & \uline{0.9868} & \textbf{0.8554} & \textbf{0.5335} & \textbf{0.5177} & \textbf{0.8188} & \textbf{0.7798} \\
\multicolumn{1}{c}{}  &\Large~~±0.0007     & \Large~~±0.0054  & \Large~~±0.0024  &\Large~~±0.0008 & \Large~~±0.0040    &\Large~~±0.0022   &\Large~~±0.0055   &\Large~~±0.0035 &\Large~~±0.0005     &\Large~~±0.0025   & \Large~~±0.0022  &\Large~~±0.0009 &\Large~~±0.0040     & \Large~~±0.0046  &\Large~~±0.0042   &\Large~~±0.0041 & \multicolumn{1}{c}{}  \\
\bottomrule
\end{tabular}
        }
    }
    \vspace{-1em}
    \label{tb:main}
\end{table*}

We conducted comparative experiments with ten state-of-the-art time series anomaly detection methods: LSTMAD~\cite{lstmad}, FCVAE~\cite{fcvae}, SRCNN~\cite{srcnn}, FITS~\cite{xu2023fits}, TimesNet~\cite{timesnet}, OFA~\cite{ofa}, TranAD~\cite{tuli2022tranad}, SubLOF~\cite{lof}, Anomaly Transformer~\cite{anomalytransformer} (abbreviated as AnoTrans in the tables), KAN~\cite{kan} and SAND~\cite{sand}. Detailed descriptions of these methods can be found in~\Cref{ap:baseline}. \deleted[id=zq]{To ensure the reliability of our experimental results, we directly adopted the hyperparameter settings reported in the original baseline papers for datasets used therein.} For datasets not featured in the baseline literature, we meticulously tuned hyperparameters via grid search to optimize the performance of the baseline method on the respective evaluation metrics.

\subsubsection{Evaluation metrics} \label{eval:metrics}

In practical applications, operations teams are less concerned with point-wise anomalies (i.e., whether individual data points are classified as anomalous) and more focused on detecting sustained anomalous segments within time series data. Furthermore, due to the potential impact of such segments, early identification is crucial. Previous work~\cite{donut} proposed the \textbf{Best F1} metric, which iterates over all thresholds and applies a point adjustment strategy to calculate the F1 score. However, it has been criticized for performance inflation~\cite{suite_tods,UCR}. 

\begin{figure}[t]
    \vspace{1em}
    \setlength{\abovecaptionskip}{-2em}
    \setlength{\belowcaptionskip}{-1.5em}
    \begin{center}  
        \includegraphics[width=1\linewidth]{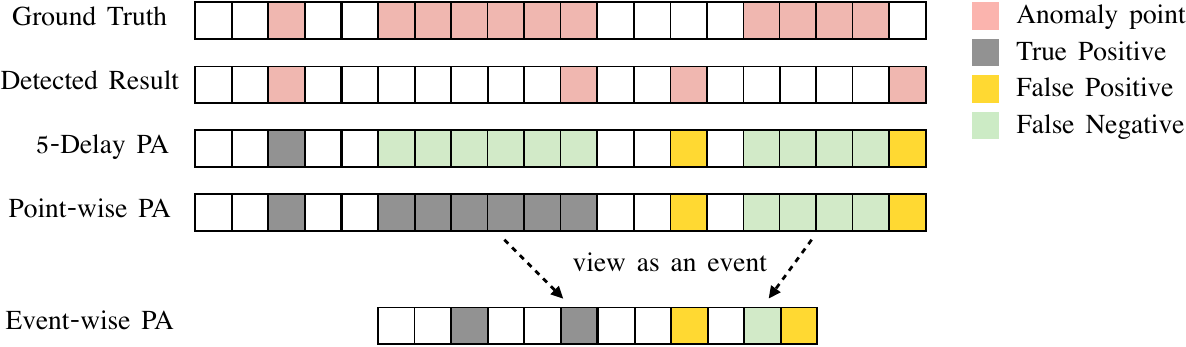}
    \end{center}
    \caption{Illustration of the adjustment strategy. Point-wise PA gives an inflated score when some anomaly segments persist for a long duration. Event-wise PA treats each anomaly segment as an event, completely disregarding the length of the anomaly segment. $k$-delay PA considers only anomalies detected within the first $k$ points after the anomaly onset, treating any detected later as undetected.}
    \label{fig:metric}
    \vspace{-3em}
\end{figure}

To address this, we also adopt \textbf{Delay F1}~\cite{srcnn} and \textbf{Event F1}. Delay F1 is similar to Best F1 but uses a delay point adjustment strategy. As shown in~\Cref{fig:metric}, the second anomaly was missed because the detection delay exceeded the threshold of five time intervals. In all experiments, a delay threshold of five was used across all datasets. Event F1, on the other hand, treats anomalies of varying lengths as anomalies with a length of 1, minimizing performance inflation caused by excessively long anomalous segments. For the sake of convenience, unless otherwise stated, we use Event F1 as the primary metric, as it is more alignment with the need for real-time anomaly detection in real-world situations.

\subsection{RQ1. Performance and Efficiency Comparison}
We present a comprehensive evaluation of \name across multiple time series anomaly detection (TSAD) experiments, with results summarized in~\Cref{tb:main}. Our analysis focuses on three key dimensions: detection accuracy, model efficiency, and computational requirements.
\begin{table}[t]
    \centering
    \addtolength{\tabcolsep}{-2pt}
    \caption{Efficiency comparison on UCR dataset.}
    \resizebox{0.9\linewidth}{!}{
        \renewcommand\arraystretch{0.95}
        \setlength{\tabcolsep}{1.5mm}{
            \centering
\begin{tabular}{lrrrr} 
\toprule
\textbf{Method} & \textbf{GPU Time} & \textbf{CPU Time} & \textbf{Parameters} & $\mathsf{F1_{e}}$ \\ 
\hline
SAND & - & 5637s & - & 0.5108 \\
SubLOF & - & 299s & - & 0.4772 \\
OFA & 220s & 3087s & 81.920 M & 0.3176 \\
AnoTrans & 201s & 1152s & 4.752 M & 0.1696 \\
FCVAE & 2327s & 1743s & 1.414 M & 0.3812 \\
TimesNet & 182s & 259s & 73,449 & 0.1805 \\
LSTMAD & 73s & 267s & 10,421 & 0.3482 \\
KAN & 66s & \uline{34s} & 1,360 & \uline{0.4120} \\
FITS & \textbf{32s} & \textbf{17s} & 624 & 0.2926 \\
TranAD & 113s & 62s & \uline{369} & 0.1840 \\ 
\hline
KAN-AD & \uline{42s} & 36s & \textbf{274} & \textbf{0.5335} \\
\bottomrule
\end{tabular}
        }
    }
    \label{tb:time}
    \vspace{-2em}
\end{table}
Across diverse experimental settings, KAN-AD demonstrates consistent and robust performance advantages. In the TODS dataset, where training data contains a substantial proportion of anomalies, \name significantly outperforms SOTA by 27\% on Event F1, highlighting its robust learning capabilities in handling noisy training data. For datasets exhibiting strong periodic characteristics (WSD and KPI), \name achieves comparable or superior performance relative to state-of-the-art approaches. Even in the challenging UCR dataset scenario, where the training set lacks anomaly samples and contains significant periodic variations, \name effectively captures normal patterns, whereas baseline methods show reduced effectiveness in pattern recognition. Quantitatively, \name achieves more than a 15\% improvement in average Event F1 score compared to existing state-of-the-art methods.

The computational efficiency analysis, presented in~\Cref{tb:time}, reveals another distinctive advantage of \namenospace. We note that several baseline methods are excluded from this comparison due to implementation constraints: SAND's CPU-only execution requirement and SubLOF's limited multi-core utilization capabilities preclude fair comparison in modern hardware acceleration contexts. Among the other models, we observe a wide spectrum of model complexities, with parameter counts ranging from millions to hundreds. Large-scale models like OFA utilize 81.92M parameters, while established approaches such as Anomaly Transformer, FCAVE, and TimesNet employ between 73k and 4.75M parameters. In contrast, \name achieves competitive performance with remarkable efficiency, requiring only 274 parameters, a 25\% reduction compared to TranAD, the next most compact model in our comparison.

These empirical findings underscore \namenospace's exceptional efficiency-performance in TSAD tasks. By achieving state-of-the-art or near state-of-the-art performance while significantly reducing the parameter footprint, \name demonstrates the effectiveness of our design principles in creating efficient, practical solutions. This combination of high detection accuracy and minimal computational requirements positions \name as an ideal choice for resource-constrained or cost-sensitive deployment scenarios, offering a compelling balance between model complexity and detection capabilities.

\subsubsection{Case Study} 
\begin{figure}[t]
\vspace{1em}
    \setlength{\abovecaptionskip}{0.5em}
    \setlength{\belowcaptionskip}{-1em}
    \centering
    \includegraphics[width=1\linewidth]{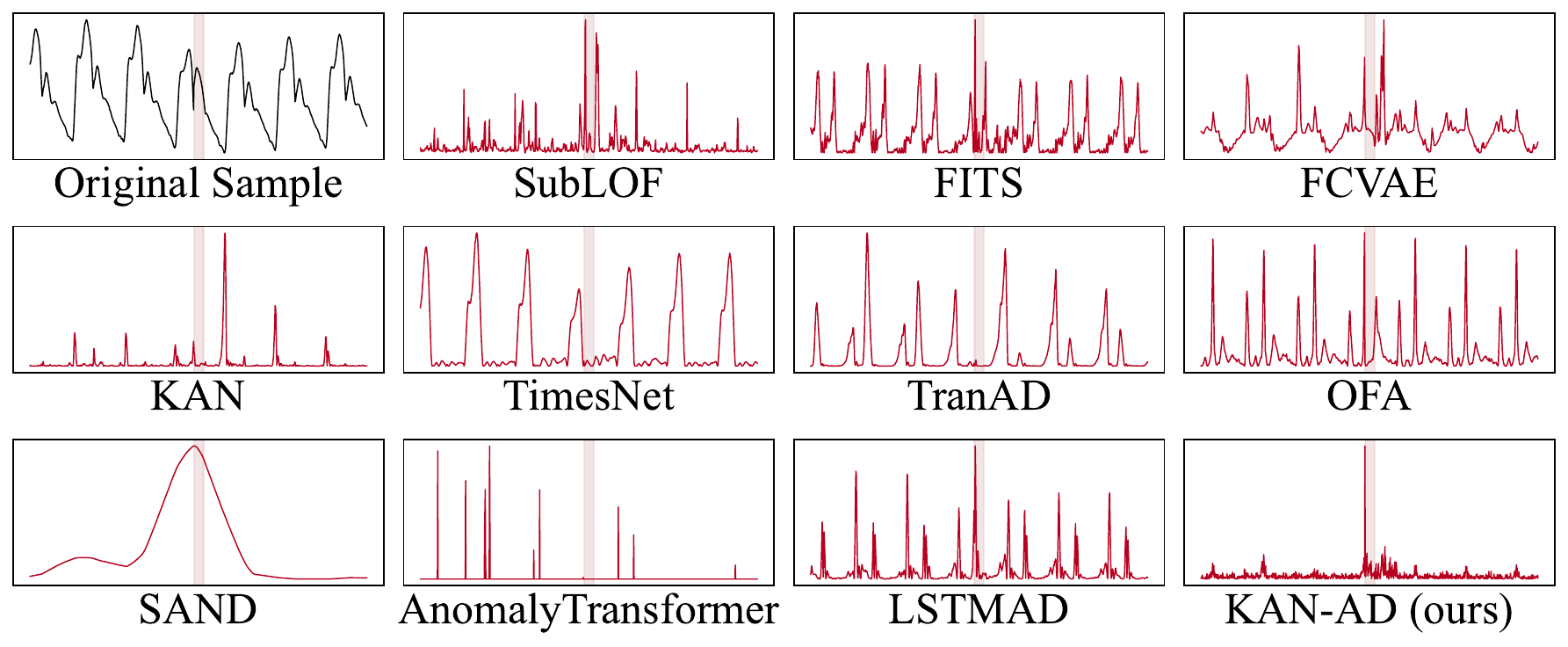}
    \vspace{-2em}
    \caption{Case study on UCR InternalBleeding10. The black curve represents the original sample, the red curve represents the anomaly scores provided by the method, and the true anomaly segments are highlighted in pink.}
    \label{fig:case}
    \vspace{-2em}
\end{figure}

\begin{figure}[t]
    \vspace{1em}
    \setlength{\abovecaptionskip}{0.5em}
    \setlength{\belowcaptionskip}{-2em}
    \centering
    \begin{minipage}[t]{0.42\linewidth}
        \centering
        \includegraphics[width=0.98\linewidth]{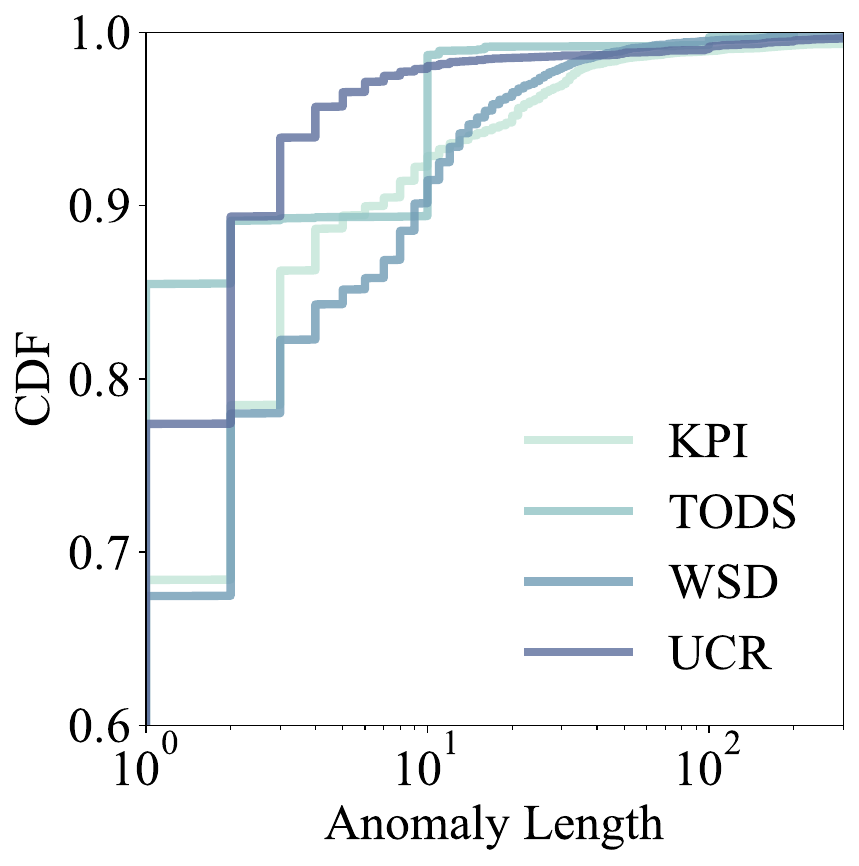}
        \caption{Anomalous lengths distribution.}
        \label{fig:dataset-cdf}
    \end{minipage}
    \hfill
    \begin{minipage}[t]{0.55\linewidth}
        \centering
        \resizebox{1\linewidth}{!}{\usetikzlibrary{calc}
\pgfplotsset{compat=1.18}
\pgfkeys{/tikz/.cd,
cube top color/.store in=\CubeTopColor,
cube top color=blue!10,
cube front color/.store in=\CubeFrontColor,
cube front color=blue!10,
cube side color/.store in=\CubeSideColor,
cube side color=blue!10,
3d cube color/.code={\colorlet{mycolor}{#1}%


\tikzset{cube top color=mycolor!90,cube front color=mycolor!60,%
cube side color=mycolor!70,draw=mycolor}}
}
\makeatletter
\pgfdeclareplotmark{half cube*}
                {%
                        \pgfplots@cube@gethalf@x
                        \let\pgfplots@cube@halfx=\pgfmathresult
                        \pgfplots@cube@gethalf@y
                        \let\pgfplots@cube@halfy=\pgfmathresult
                        \pgfplots@cube@gethalf@z
                        \let\pgfplots@cube@halfz=\pgfmathresult
                        \pgfmathparse{0*\pgfplots@cube@halfz}%
                        \let\pgfplots@cube@topz=\pgfmathresult
                        \pgfmathparse{-1*\pgfplots@cube@halfz}%
                        \let\pgfplots@cube@bottomz=\pgfmathresult
                        \pgfplotsifaxissurfaceisforeground{0vv}{%
                                \pgfsetfillcolor{\CubeFrontColor}
                                \pgfpathmoveto{\pgfplotsqpointxyz{-\pgfplots@cube@halfx}{-\pgfplots@cube@halfy}{\pgfplots@cube@bottomz}}%
                                \pgfpathlineto{\pgfplotsqpointxyz{-\pgfplots@cube@halfx}{-\pgfplots@cube@halfy}{\pgfplots@cube@topz}}%
                                \pgfpathlineto{\pgfplotsqpointxyz{-\pgfplots@cube@halfx}{ \pgfplots@cube@halfy}{\pgfplots@cube@topz}}%
                                \pgfpathlineto{\pgfplotsqpointxyz{-\pgfplots@cube@halfx}{ \pgfplots@cube@halfy}{\pgfplots@cube@bottomz}}%
                                \pgfpathclose
                                \pgfusepathqfillstroke
                        }{%
                                \pgfsetfillcolor{\CubeFrontColor}
                                \pgfpathmoveto{\pgfplotsqpointxyz{ \pgfplots@cube@halfx}{-\pgfplots@cube@halfy}{\pgfplots@cube@bottomz}}%
                                \pgfpathlineto{\pgfplotsqpointxyz{ \pgfplots@cube@halfx}{-\pgfplots@cube@halfy}{\pgfplots@cube@topz}}%
                                \pgfpathlineto{\pgfplotsqpointxyz{ \pgfplots@cube@halfx}{ \pgfplots@cube@halfy}{\pgfplots@cube@topz}}%
                                \pgfpathlineto{\pgfplotsqpointxyz{ \pgfplots@cube@halfx}{ \pgfplots@cube@halfy}{\pgfplots@cube@bottomz}}%
                                \pgfpathclose
                                \pgfusepathqfillstroke
                        }%
                        \pgfplotsifaxissurfaceisforeground{v0v}{%
                                \pgfsetfillcolor{\CubeSideColor}
                                \pgfpathmoveto{\pgfplotsqpointxyz{-\pgfplots@cube@halfx}{-\pgfplots@cube@halfy}{\pgfplots@cube@bottomz}}%
                                \pgfpathlineto{\pgfplotsqpointxyz{-\pgfplots@cube@halfx}{-\pgfplots@cube@halfy}{\pgfplots@cube@topz}}%
                                \pgfpathlineto{\pgfplotsqpointxyz{ \pgfplots@cube@halfx}{-\pgfplots@cube@halfy}{\pgfplots@cube@topz}}%
                                \pgfpathlineto{\pgfplotsqpointxyz{ \pgfplots@cube@halfx}{-\pgfplots@cube@halfy}{\pgfplots@cube@bottomz}}%
                                \pgfpathclose
                                \pgfusepathqfillstroke
                        }{%
                                \pgfsetfillcolor{\CubeSideColor}
                                \pgfpathmoveto{\pgfplotsqpointxyz{-\pgfplots@cube@halfx}{ \pgfplots@cube@halfy}{\pgfplots@cube@bottomz}}%
                                \pgfpathlineto{\pgfplotsqpointxyz{-\pgfplots@cube@halfx}{ \pgfplots@cube@halfy}{\pgfplots@cube@topz}}%
                                \pgfpathlineto{\pgfplotsqpointxyz{ \pgfplots@cube@halfx}{ \pgfplots@cube@halfy}{\pgfplots@cube@topz}}%
                                \pgfpathlineto{\pgfplotsqpointxyz{ \pgfplots@cube@halfx}{ \pgfplots@cube@halfy}{\pgfplots@cube@bottomz}}%
                                \pgfpathclose
                                \pgfusepathqfillstroke
                        }%
                        \pgfplotsifaxissurfaceisforeground{vv0}{%
                                \pgfsetfillcolor{\CubeTopColor}
                                \pgfpathmoveto{\pgfplotsqpointxyz{-\pgfplots@cube@halfx}{-\pgfplots@cube@halfy}{\pgfplots@cube@bottomz}}%
                                \pgfpathlineto{\pgfplotsqpointxyz{-\pgfplots@cube@halfx}{ \pgfplots@cube@halfy}{\pgfplots@cube@bottomz}}%
                                \pgfpathlineto{\pgfplotsqpointxyz{ \pgfplots@cube@halfx}{ \pgfplots@cube@halfy}{\pgfplots@cube@bottomz}}%
                                \pgfpathlineto{\pgfplotsqpointxyz{ \pgfplots@cube@halfx}{-\pgfplots@cube@halfy}{\pgfplots@cube@bottomz}}%
                                \pgfpathclose
                                \pgfusepathqfillstroke
                        }{%
                                \pgfsetfillcolor{\CubeTopColor}
                                \pgfpathmoveto{\pgfplotsqpointxyz{-\pgfplots@cube@halfx}{-\pgfplots@cube@halfy}{\pgfplots@cube@topz}}%
                                \pgfpathlineto{\pgfplotsqpointxyz{-\pgfplots@cube@halfx}{ \pgfplots@cube@halfy}{\pgfplots@cube@topz}}%
                                \pgfpathlineto{\pgfplotsqpointxyz{ \pgfplots@cube@halfx}{ \pgfplots@cube@halfy}{\pgfplots@cube@topz}}%
                                \pgfpathlineto{\pgfplotsqpointxyz{ \pgfplots@cube@halfx}{-\pgfplots@cube@halfy}{\pgfplots@cube@topz}}%
                                \pgfpathclose
                                \pgfusepathqfillstroke
                        }%
            }
\makeatother
\begin{tikzpicture}[scale=0.75, transform shape]
\definecolor{color1}{rgb}{0.76110584, 0.89771305, 0.84543495}
\definecolor{color2}{rgb}{0.56920577, 0.76982577, 0.77131702}
\definecolor{color3}{rgb}{0.44001013, 0.60735469, 0.71168273}
\definecolor{color4}{rgb}{0.36445067, 0.43100839, 0.61584039}
\definecolor{color5}{rgb}{0.29790922, 0.2562451 , 0.45126194}

\pgfmathsetmacro{\gconv}{3000}
\begin{axis}[
view={120}{30},
width=10cm,
height=10cm,
grid=major,
xmin=0,xmax=6,
ymin=0,ymax=6,
zmin=0.75,zmax=0.84,
xtick={1,2,3,4,5},
xticklabels={16,32,64,96,128},
ytick={1,2,3,4,5},
yticklabels={1,2,4,8,16},
ylabel={$N$},
xlabel={$T$},
zlabel={$AUPRC$},
tick label style={font=\Large},
label style={font=\Large},
x dir=reverse,
]
\path let \p1=($(axis cs:0,0,1)-(axis cs:0,0,0)$) in 
\pgfextra{\pgfmathsetmacro{\conv}{2*\y1}
\ifx\gconv\conv
\typeout{z-scale\space good!}
\else
\typeout{Kindly\space consider\space setting\space the\space 
        prefactor\space of\space z\space to\space \conv}
\fi     
        };  

\pgfplotsset{3d bars/.style={only marks,scatter,mark=half cube*,mark size=0.45cm, 
3d cube color=#1,point meta=0,
visualization depends on={\gconv*(z-0.75) \as \myz}, 
scatter/@pre marker code/.append style={/pgfplots/cube/size z=\myz},draw=#1}}

\addplot3[
    3d bars=color5
]
coordinates {(5,1,0.8119)(5,2,0.8186)(5,3,0.8250)(5,4,0.8253)(5,5,0.8222)};
\addplot3[
    3d bars=color4
]
coordinates {(4,1,0.8034)(4,2,0.8221)(4,3,0.8226)(4,4,0.8195)(4,5,0.8216)};
\addplot3[
    3d bars=color3
]
coordinates {(3,1,0.7929)(3,2,0.8172)(3,3,0.8213)(3,4,0.8214)(3,5,0.8180)};
\addplot3[
    3d bars=color2
]
coordinates {(2,1,0.7734)(2,2,0.7990)(2,3,0.8141)(2,4,0.8122)(2,5,0.8088)};
\addplot3[
    3d bars=color1
]
coordinates {(1,1,0.7660)(1,2,0.7844)(1,3,0.7905)(1,4,0.8004)(1,5,0.7801)};

\end{axis}
\end{tikzpicture}}
        \caption{Model performance under different hyperparameters.}
        \label{fig:hp}
    \end{minipage} 
    \vspace{-2em}
\end{figure}

We analyzed anomaly detection performance on UCR dataset samples to illustrate how various methods respond to identical anomalies, as shown in~\Cref{fig:case}. The selected sample displayed pattern anomalies, marked by significant deviations from typical behavior. Both TranAD and TimesNet exhibit difficulty establishing normal patterns. Minor variations among normal samples across cycles lead to periodic false alarms during normal segments, consistent with our observations in~\Cref{fig:noise-sensitive}. Among the methods listed, while OFA, LSTMAD, SubLOF, and FITS can detect anomalies, their high anomaly scores during normal segments indicate excessive sensitivity to minor fluctuations in normal data. In contrast, \name excels in identifying anomalies while maintaining minimal anomaly scores during normal segments. 

\subsection{RQ2. Hyperparameter sensitivity} \label{eval:hp}

The \name model incorporates two key hyperparameters: the number of terms in univariate functions $N$ and the window size $T$. To investigate the ultimate impact of these parameters on model performance, we conducted experiments on the UCR dataset while holding all other parameters constant. As findings summarized in~\Cref{fig:hp}, a larger window size facilitates more accurate learning of normal patterns when $N$ is fixed, leading to improved performance. When $T$ is fixed, insufficient univariate functions limit \namenospace's expressive power, while excessive $N$ can lead to overfitting. Overall, \name achieved its best performance with $T=96$ and $N=2$. Notably, even with suboptimal hyperparameter settings like $T=16$ and $N=1$, we surpassed SOTA methods on the UCR dataset.

\subsection{RQ3. Ablation Studies} \label{eval:ablation}

\begin{figure*}[htp]
    \setlength{\abovecaptionskip}{-1em}
    \setlength{\belowcaptionskip}{-1em}
    \centering
    \begin{minipage}{0.32\textwidth}
        \centering
        \includegraphics[width=\linewidth]{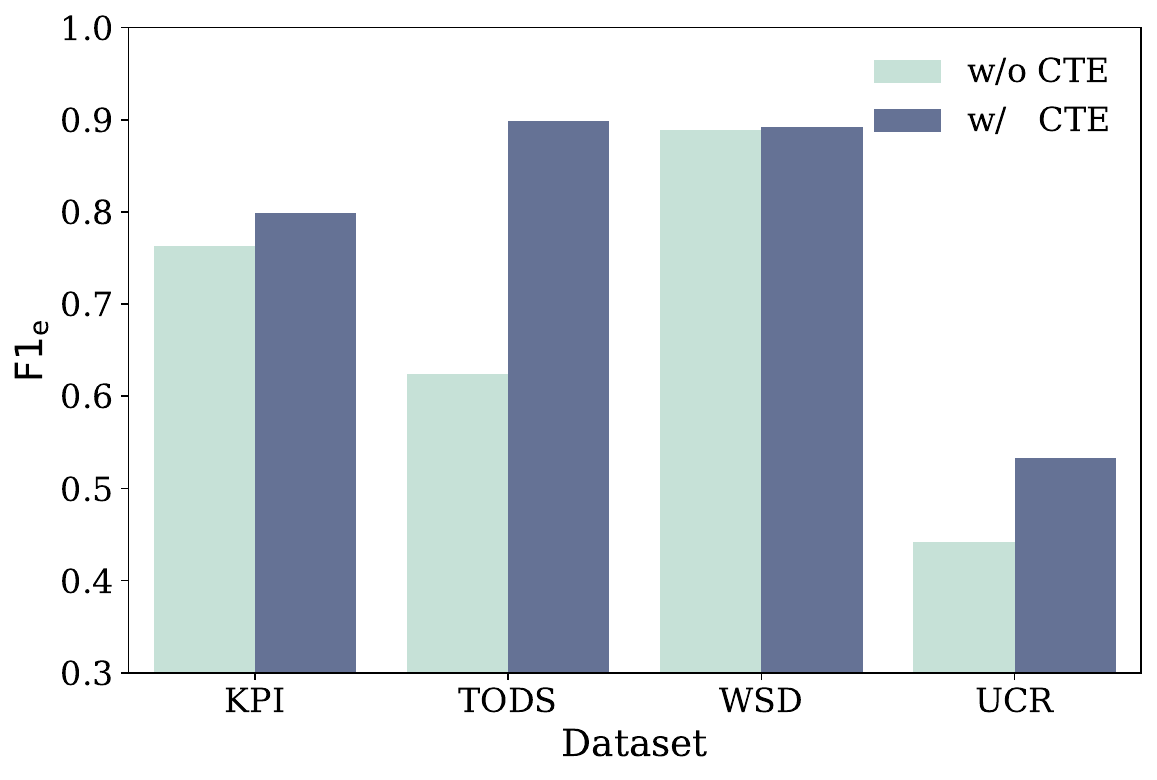}
        \caption{Model performance under \\different preprocessing.}
        \label{fig:diff}
    \end{minipage}
    \hspace{0.1em}
    \centering
    \begin{minipage}{0.32\textwidth}
        \centering
        \includegraphics[width=\linewidth]{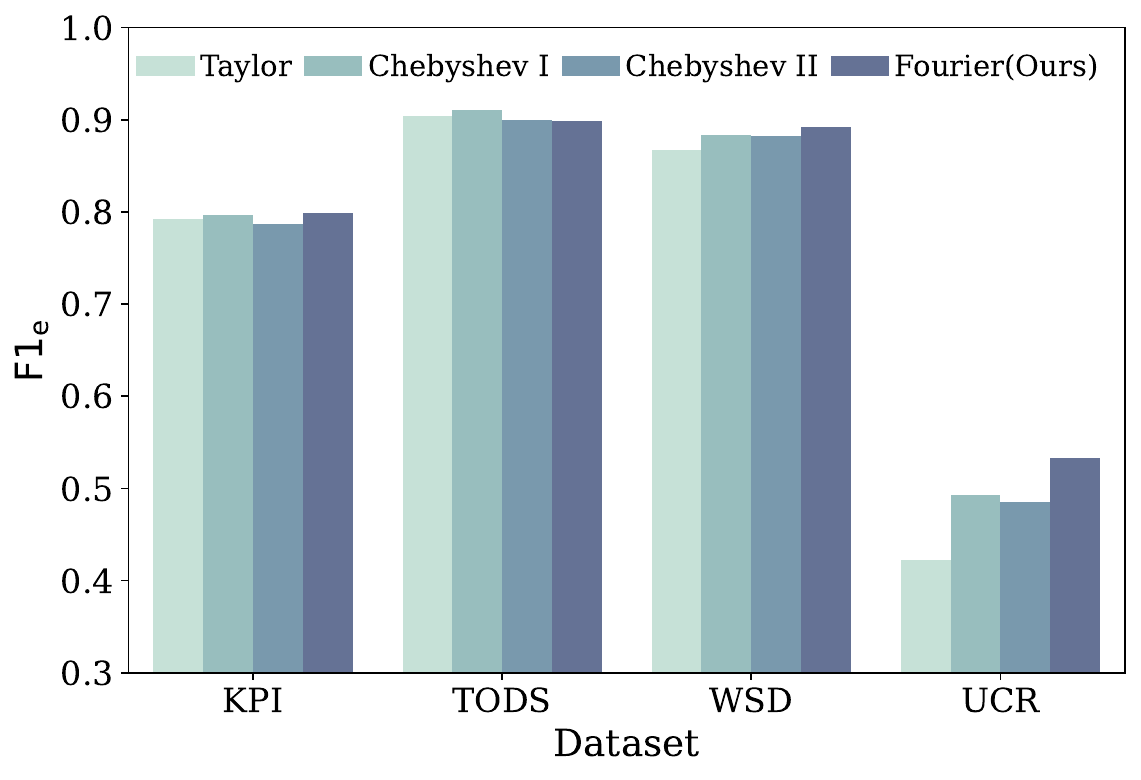}
        \caption{Model performance under \\different univariate function.}
        \label{fig:basic}
    \end{minipage}
    \hspace{0.1em}
    \begin{minipage}{0.32\textwidth}
        \centering
        \includegraphics[width=\linewidth]{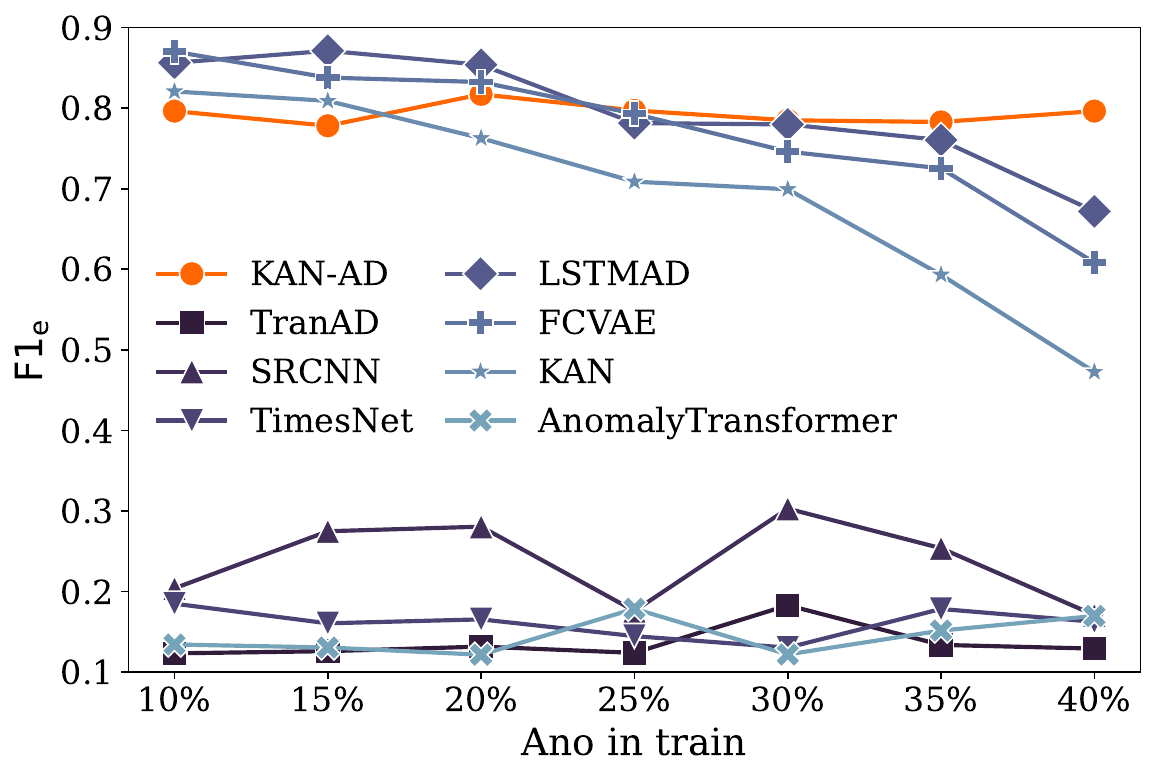}
        \caption{Model performance under \\different anomaly ratios in training.}
        \label{fig:noise-ratio}
    \end{minipage}
    \vspace{-1em}
\end{figure*}


In this section, we investigated the impact of constant term elimination modules, different univariate function selections on algorithm performance and the influence of the \mname~mechanism.

\subsubsection{Constant term elimination module} 
We employed a constant term elimination (CTE) module during data preprocessing to mitigate the influence of the offset term $A_0$ in~\Cref{agn:fourier-formula}. Further experiments were conducted across all datasets to evaluate the impact of incorporating CTE within the preprocessing pipeline. As presented in~\Cref{fig:diff}, the impact of CTE varies across datasets, reflecting inherent data characteristics. For datasets with pronounced periodicity or strong temporal stability (e.g., WSD), the benefits of CTE are less apparent. Conversely, for datasets exhibiting larger value fluctuations or trends (e.g., KPI, TODS and UCR), CTE yields significant improvements.

\subsubsection{Selection of univariate functions}

\begin{table}[t]
    \centering
    \caption{Commonly used univariate functions for time series approximation.}
        \renewcommand\arraystretch{0.85}
        \setlength{\tabcolsep}{3mm}{
            \centering
\begin{tabular}{lr} 
\toprule
\textbf{ Name } & \textbf{$\Phi_n(x)$}\\ 
\midrule
Taylor Series & $x^n$ \\
Fourier Series & $\cos(nx)+\sin(nx)$ \\
Chebyshev Polynomial \textup{\uppercase\expandafter{\romannumeral 1}} & $\cos(n \arccos(x))$ \\
Chebyshev Polynomial \textup{\uppercase\expandafter{\romannumeral 2}} & $\frac{\sin((n+1)\arccos(x))}{\sin(\arccos(x))}$ \\
\bottomrule
\end{tabular}
        }
    \label{tab:approximation}
    \vspace{-2em}
\end{table}

\replaced[id=zq]{To assess the impact of different univariate functions on model performance,}{To comprehensively assess the ability of \namenospace,} we conducted experiments using common univariate functions listed in~\Cref{tab:approximation}. In our implementations, due to varying input range requirements across univariate functions, appropriate normalization techniques are employed. Specifically, min-max scaling to the range $x\in[-1, 1]$ was utilized for both types of Chebyshev polynomials, while z-score was employed for Taylor series and Fourier series. The performance of all four univariate functions was compared using the same configuration. As results presented in~\Cref{fig:basic}, Fourier series consistently achieved the top two performance across all datasets. In contrast, Taylor series exhibited persistent bias due to non-zero function values in most cases, hindering optimal model performance. The objective of both types of Chebyshev polynomials is to minimize the maximum error, which potentially conflicts with anomaly detection methods that minimize mean squared prediction error, thus leading to suboptimal performance.

\subsection{RQ4. Robustness to Anomalous Data}

To evaluate \namenospace's robustness to anomalies in the training set, we conducted additional experiments using synthetic datasets constructed in accordance with the TODS dataset generation methodology. We synthesized test datasets containing local peaks and drops anomalies, and progressively increased the proportion of these anomalies in the initially anomaly-free training set. As illustrated in~\Cref{fig:noise-ratio}, \name demonstrates stable performance across all anomaly ratios. Popular methods such as LSTMAD, perform well at lower anomaly ratios but experience a significant decline as the ratio increases. Other approaches, like TranAD, fail to achieve optimal performance due to overfitting to fine-grained structures within the time series.


\subsection{Ablation on function deconstruction mechanism} \label{ap:decomp}
To investigate the impact of the function deconstruction mechanism, we compared the model's detection capabilities under different univariate function combination strategies. 
For clarity, the specific definitions are provided in~\Cref{agn:feature-set}. As the results presented in~\Cref{tab:feature}, the model's detection performance exhibited a notable improvement with an increasing number of univariate functions. Both Fourier series and cosine waves outperformed the raw input data, likely due to their smoother representations compared to the original signal, enabling higher detection accuracy. The combination of different features, particularly those involving Fourier series and cosine waves, resulted in significant performance gains as the feature count increased. Ultimately, \name achieved optimal detection performance by integrating all features. It is worth noting that even the variant of \name utilizing only the raw time series X outperforms KAN, clearly demonstrating the advantage of Fourier series over the use of spline functions for optimizing univariate functions.

\begin{table}[t]
    \centering
    \addtolength{\tabcolsep}{3pt}
    \caption{Model performance on UCR dataset under different function deconstruction strategies.}
    \renewcommand\arraystretch{0.90}
    \begin{tabular}{lrrr}
\toprule
 \textbf{Variation}  & \textbf{$\mathsf{F1_{e}}$} & \textbf{$\mathsf{F1_{d}}$} & \textbf{ $\mathsf{AUPRC}$ } \\
\hline
~\namenospace & 0.5335   & 0.5177   & 0.8188  \\ 
\small~~~~$\mathrm{w/o~X}$ & 0.5153   & 0.4974   & 0.8066  \\ 
\small~~~~$\mathrm{w/o~P}$ & 0.5081   & 0.4810    & 0.8007  \\ 
\small~~~~$\mathrm{w/o~S}$ & 0.5056   & 0.5113   & 0.7998  \\ 
\small~~~~$\mathrm{w/o~X\&P} $  & 0.4737   & 0.4583   & 0.7872  \\ 
\small~~~~$\mathrm{w/o~X\&S} $  & 0.4698   & 0.4610    & 0.7767  \\ 
\small~~~~$\mathrm{w/o~S\&P}$   & 0.4561   & 0.4637   & 0.7595  \\ 
\bottomrule
\end{tabular}
    \label{tab:feature}
\end{table}

\subsection{Performance on Multivariate Time Series}
\begin{table*}[htb]
    \centering
    \caption{Best F1 and parameter counts for multivariate time series anomaly detection. Best and second best results are in \textbf{bold} and  \uline{underline}.}
    \begin{tabular}{lccccccr}
\hline
\textbf{Methods}        & \textbf{SMD}   & \textbf{MSL}   & \textbf{SMAP}  & \textbf{SWaT}  & \textbf{PSM}   & \textbf{Avg F1} & \textbf{Parameters@MSL} \\ \hline
Informer~\cite{zhou2021informer}              & 0.8165         & 0.8406         & 0.6992         & 0.8143         & 0.7710         & 0.7883          & 504,174                 \\
Anomaly Transformer~\cite{anomalytransformer}   & \uline{0.8549}& 0.8331         & 0.7118         & 0.8310         & 0.7940         & 0.8050          & 4,863,055               \\
DLinear~\cite{zeng2023dlinear}               & 0.7710         & \uline{0.8488}& 0.6926         & 0.8752         & 0.9355         & 0.8246          & 20,200                  \\
Autoformer~\cite{wu2021autoformer}            & 0.8511         & 0.7905         & 0.7112         & 0.9274         & 0.9329         & 0.8426          & 325,431                 \\
FEDformer~\cite{zhou2022fedformer}             & 0.8508         & 0.7857         & 0.7076         & 0.9319         & 0.9723         & 0.8497          & 1,119,982               \\
TimesNet~\cite{timesnet}              & 0.8462         & 0.8180         & 0.6950         & 0.9300         & \uline{0.9738}& 0.8526          & 75,223                  \\
UniTS~\cite{gao2024units}                 & \textbf{0.8809}& 0.8346         & \uline{0.8380}& \uline{0.9326}& \textbf{0.9743}& \uline{0.8921} & 8,066,376               \\ \hline
\textbf{KAN-AD (ours)}  & 0.8429         & \textbf{0.8501}& \textbf{0.9450}& \textbf{0.9350}& 0.9650         & \textbf{0.9076} & \textbf{4,491}          \\ \hline
\end{tabular}

    \label{tab:mts_results}
\end{table*}

To extend \namenospace's application to the multivariate time series (MTS) scenario, we adopt a channel-independent approach. Specifically, an MTS input with the shape \texttt{(batch\_size, window\_length, n\_features)} is reshaped into \texttt{(batch\_size * n\_features, window\_length)}. Each of the \texttt{n\_features} channels is thus treated as an independent univariate time series instance. \name is then applied to these individual series. This channel-independent strategy has proven effective~\cite{patchtst}. By adopting a similar principle, \name can leverage its robust univariate modeling capabilities across all channels of an MTS dataset. The model is trained on the collection of these reshaped univariate instances, allowing it to learn generalized normal patterns.

We implemented MTS versions of \name in popular time series library~\cite{THUML} and evaluated them on the common SMD~\cite{smd}, MSL~\cite{msl}, SMAP~\cite{smap}, SWaT~\cite{swat}, and PSM~\cite{psm} datasets. Our evaluation metric uses the Best F1 score which is consistent with the baseline methods. We introduce these datasets and baseline methods in detail in the~\Cref{ap:mts}. As detailed in~\Cref{tab:mts_results}, KAN-AD achieves the highest average Best F1 score of 0.9076, across all five benchmark datasets, outperforming all listed SOTA methods. A significant advantage of KAN-AD is its exceptional parameter efficiency. With only \textit{4,491} trainable parameters (measured on MSL), KAN-AD utilizes substantially fewer parameters than all other compared methods.

\section{Related Work} \label{related}
\textbf{Time Series Forecasting Methods}: These methods can be categorized into prediction-based and reconstruction-based methods, both aiming to identify deviations from normal patterns through temporal analysis.
Prediction-based methods, like FITS~\cite{xu2023fits} achieves efficient detection through frequency domain analysis with minimal parameters, while LSTMAD~\cite{lstmad} leverages LSTM networks~\cite{lstm} to capture complex temporal dependencies.
Reconstruction-based approaches, like Donut~\cite{donut} focus on time series denoising, while FCVAE~\cite{fcvae} enhances the VAE~\cite{vae} framework by incorporating frequency domain information. Recent advances in Transformer architectures have further strengthened reconstruction capabilities: TranAD~\cite{tuli2022tranad} employs adversarial learning for robust pattern capture, while OFA~\cite{ofa} leverages GPT-2~\cite{gpt2} for modeling complex temporal dependencies.

\noindent\textbf{Pattern Change Detection Methods}: These approaches identify anomalies through comparative analysis of current and historical patterns. Early methods, like SubLOF~\cite{lof} quantify pattern variations using window-based distance metrics. SAND employs temporal shape-based clustering to distinguish anomalous patterns. Recent advances, exemplified by TriAD~\cite{triad}, leverage multi-domain contrastive learning frameworks, demonstrating superior performance on UCR datasets.
\section{Conclusion} \label{conclusion}

Training time series anomaly detection models with datasets containing anomalies is essential for deployment in production environments. Existing algorithms often rely on {carefully selected features and} complex architectures to achieve minor accuracy gains, neglecting robustness during training. This paper introduces \namenospace, a robust anomaly detection model rooted in the Kolmogorov–Arnold representation theorem. \name transforms the prediction of time points into the estimation of coefficients of Fourier series, achieving strong performance with few parameters, significantly reducing costs while enhancing robustness to outliers. \name includes a constant term elimination module to address temporal trends and leverages frequency domain information for better performance. \replaced[id=zq]{\name surpasses the SOTA model across four public datasets with a 15\% improvement in average Event F1 score, simultaneously achieving an 80\% reduction in parameter count and 50\% faster inference speed compared to vanilla KAN.}{Compared to the SOTA model across four public datasets, \name achieves a 15\% improvement in average Event F1 score, while reducing the parameter count by 80\% and accelerating inference speed by 50\% compared to vanilla KAN.} With \namenospace, a promising research direction is to explore whether normal patterns in time series can be represented more efficiently by leveraging additional data.

\section*{Acknowledgments}

This work was partially funded by the National Key Research and Development Program of China (No.2022YFB2901800), the National Natural Science Foundation of China (62202445), the National Natural Science Foundation of China-Research Grants Council (RGC) Joint Research Scheme (62321166652), and the National Natural Science Foundation of China (Grant No. W2412136).

\section*{Impact Statement}

This paper presents work whose goal is to advance the field of 
Machine Learning. There are many potential societal consequences 
of our work, none which we feel must be specifically highlighted here.

\bibliographystyle{icml2025}
\bibliography{main}

\newpage
\appendix
\onecolumn
\section{Datasets and Baseliens on Univariate Time Series}
\subsection{Datasets} \label{ap:dataset}
We selected four datasets from diverse domains, with samples originating from:
\begin{itemize}
    \item \textbf{KPI}~\cite{KPI}: This dataset comprises service metrics collected from five major Internet companies: Sogou, eBay, Baidu, Tencent, and Alibaba. The data points are primarily recorded every 1-2 minutes, with some sections exhibiting a 5-minute interval.
    \item \textbf{TODS}~\cite{suite_tods}: TODS comprises artificially created time series, each designed to present specific types of anomalies. Its excellent interpretability and carefully constructed data distributions make it suitable for in-depth case studies.
    \item \textbf{WSD}~\cite{WSD}: This dataset consists of web server metrics collected from three companies providing large-scale web services: Baidu, Sogou, and eBay.
    \item \textbf{UCR}~\cite{UCR}: This archive contains data from multiple domains with a single anomalous segment on each time series. In addition to real anomalies, UCR also includes synthetic but highly plausible anomalies.
    \item \textbf{NAB}~\cite{NAB}: We utilize the Twitter dataset within the NAB, which was published by Numenta and comprises various time series data from machine, environmental, financial, and other domains.
    \item \textbf{Yahoo}~\cite{yahoo}: Developed by Yahoo Labs, this dataset encompasses both real and synthetically generated data. The synthetic portion features fluctuating trends, noise, and seasonal patterns, while the real data encapsulates metrics from various Yahoo services.
\end{itemize}

\subsection{Baselines} \label{ap:baseline}
We selected the following baseline approaches to further elaborate on the performance differences between \name and SOTA methods:

\begin{itemize}
    \item \textbf{SubLOF}~\cite{lof} represents traditional outlier detection techniques based on distance metrics.
    \item \textbf{SRCNN}~\cite{srcnn} is a supervised approach reliant on high-quality labeled data.
    \item \textbf{LSTMAD}~\cite{lstmad} leverages Long Short-Term Memory (LSTM) networks~\cite{lstm} for deep learning-based anomaly detection.
    \item \textbf{FITS}~\cite{xu2023fits} achieves parameter-efficient anomaly detection by upsampling frequency domain information using a low-pass filter and simple linear layers.
    \item \textbf{FCVAE}~\cite{fcvae} is unsupervised reconstruction method based on Variational Autoencoder (VAE)~\cite{vae}, designed to reconstruct normal patterns.
    \item \textbf{Anomaly Transformer}~\cite{anomalytransformer} employs attention mechanism to computer the association discrepancy.
    \item \textbf{TranAD}~\cite{tuli2022tranad} incorporates the principles of adversarial learning to develop a training framework with two stages while integrating the strengths of self-attention encoders to capture the temporal dependency embedded in the time series.
    \item \textbf{SAND}~\cite{sand} utilizes a novel statistical approach based on curve shape clustering for anomaly detection in a streaming fashion.
    \item \textbf{TimesNet}~\cite{timesnet} leverages an Inception~\cite{inception}-based computer vision backbone to enhance learning capabilities. 
    \item \textbf{OFA}~\cite{ofa}, with GPT-2~\cite{gpt2} as its backbone, improves its ability to capture point-to-point dependencies. 
    \item \textbf{KAN}~\cite{kan} leverages Kolmogorov-Arnold representation theory to decompose complex learning objectives into linear combinations of univariate functions.
\end{itemize}

These baseline methods encompass a variety of anomaly detection paradigms: shape-based SAND, subsequence distance-based SubLOF, Transformer-based approaches like OFA, TranAD, and Anomaly Transformer for modeling sequence relationships, and frequency domain information enhanced methods FCVAE and FITS. 

\section{Datasets and Baselines on Multivariate Time Series} \label{ap:mts}

\subsection{Datasets}

We evaluated KAN-AD on five widely-used public benchmark datasets for multivariate time series anomaly detection:
\begin{itemize}
    \item \textbf{SMD}~\cite{smd}: A dataset collected from a large internet company, containing data from many server machines over several weeks.
    \item \textbf{MSL}~\cite{msl}: A dataset from NASA containing telemetry data from the Mars Science Laboratory rover.
    \item \textbf{SMAP}~\cite{smap}: Another NASA dataset, containing telemetry data from the SMAP satellite.
    \item \textbf{SWaT}~\cite{swat}: A dataset generated from a scaled-down real-world water treatment testbed, including normal and attack scenarios.
    \item \textbf{PSM}~\cite{psm}: A dataset from eBay, consisting of aggregated metrics from multiple application servers.
\end{itemize}

\subsection{Baselines}

We selected the following baseline approaches to further evaluate \name and SOTA methods on multivariate time series datasets:

\begin{itemize}
    \item \textbf{Informer}~\cite{zhou2021informer}: A Transformer-based model designed for long sequence time-series forecasting, featuring a ProbSparse self-attention mechanism to improve efficiency. For anomaly detection, it typically relies on reconstruction error or forecast error.
    \item \textbf{Anomaly Transformer}~\cite{anomalytransformer}: A Transformer architecture specifically tailored for time series anomaly detection, which aims to learn prior-associations and series-associations to better distinguish anomalies.
    \item \textbf{DLinear}~\cite{zeng2023dlinear}: A simple yet effective linear model that decomposes the time series into trend and remainder components, challenging the necessity of complex Transformer architectures for some forecasting tasks and adaptable for anomaly detection via reconstruction.
    \item \textbf{Autoformer}~\cite{wu2021autoformer}: A Transformer model with a novel decomposition architecture and an Auto-Correlation mechanism, designed to discover series-wise connections and improve long-term forecasting accuracy.
    \item \textbf{FEDformer}~\cite{zhou2022fedformer}: A Transformer variant that enhances performance for long sequence forecasting by employing frequency-enhanced decomposition and a mixture of expert design in the frequency domain.
    \item \textbf{TimesNet}~\cite{timesnet}: A model that transforms 1D time series into a set of 2D tensors based on identified periods and applies a 2D kernel (e.g., Inception block) to capture both intra-period and inter-period variations for general time series analysis.
    \item \textbf{UniTS}~\cite{gao2024units}: Aims to provide a unified framework for time series analysis, often leveraging large-scale pre-training on diverse datasets to build a universal representation for both univariate and multivariate time series tasks.
\end{itemize}


\end{document}